# Bimodal Distribution Removal and Genetic Algorithm in Neural Network for Breast Cancer Diagnosis

Ke Quan[1]

[1] Research School of Computer Science
Australian National University
ACT 2601 AUSTRALIA
U6646917@anu.edu.au

**Abstract.** Diagnosis of breast cancer has been well studied in the past. Multiple linear programming models have been devised to approximate the relationship between cell features and tumour malignancy. However, these models are less capable in handling non-linear correlations. Neural networks instead are powerful in processing complex non-linear correlations. It is thus certainly beneficial to approach this cancer diagnosis problem with a model based on neural network. Particularly, introducing bias to neural network training process is deemed as an important means to increase training efficiency. Out of a number of popular proposed methods for introducing artificial bias, Bimodal Distribution Removal (BDR) presents ideal efficiency improvement results and fair simplicity in implementation. However, this paper examines the effectiveness of BDR against the target cancer diagnosis classification problem and shows that BDR process in fact negatively impacts classification performance. In addition, this paper also explores genetic algorithm as an efficient tool for feature selection and produced significantly better results comparing to baseline model that without any feature selection in place.

## 1 Introduction

The diagnosis of breast cancer is critical in reducing motility rate for women at age groups with non-trial risk of breast cancer. Earlier study in Sweden shows that a 24% significant reduction of breast cancer mortality could be gained among women who are invited to diagnosis compared with those who are not [11]. Traditionally, breast cancer diagnosis was performed through by invasive surgical operation. Fine needle aspirations (FNA) provides a less intrusive means to examine the tumor by extracting a small amount of tissues from the tumor. Individual cells and contextual features could be closely examined for discovery of correlation with malignancy [12]. Traditional techniques including linear programming and decision trees were able to obtain decent prediction results. Research by Bennett in 1992 was able to reduce prediction error rate to 3.8% using two-class decision tree [13]. Later research that combines linear programming model with one separating plane in 3-D space estimated accuracy of 97.5% using 10-fold cross validation [14].

However, a few problems exist in traditional linear-programming based approaches. First of all, linear programming assumes linear relationship between dependent and independent variables, though it is certainly possible for a logical regression model to capture complex non-linear relationship, it often requires explicit search for such relationship by researchers and involves non-trial transformation of outcome variables [15]. Furthermore, particular to the breast cancer diagnosis problem, it is still unclear that how many different features are correlated with malignancy, thus certain features gathered may be entirely irrelevant to cancer diagnosis [12]. However, linear programming model is not inherently able to determine whether a feature is truly noisy or instead a contributing factor [15].

On the other hand, Artificial Neural Networks (ANN) demonstrates adequate edges in addressing these two problems. ANN as a universal classifier is able to automatically adjust connection weights to reflect complex non-linear relationships [15]. In addition, certain techniques that are popular in the field of neural network research handles feature selection pretty well. Genetic Algorithm (GA), for instance, offers an approach to select a subset of features to represent patterns to be classified with decent performance [16].

These advantages brought by ANN give rise to the motivation of devising a neural network to revisit the classification problem in breast cancer diagnosis. Particularly, we will examine if genetic algorithm is capable of filtering off cell features that are trivially contributing to tumor malignancy. Furthermore, given the fact that cell images capture from tissue extraction may contain noise and numerical presentation of features in these images may not be fully accurate, it is worthwhile to explore techniques to detect noisy patterns and exclude them from training. Bimodal distribution removal (BDR) may be an adequate candidate solution to approach this problem.

Feedforward Artificial Neural Networks (ANN) could be interpreted as statistical inferences, where the adjusting of weights falls under the class of non-parametric inference scheme [1]. Hence, challenges in the domain of non-parametric statistical inferences also apply to the training of ANN.

Bias and variance trade-off in ANN state the dilemma that lowering one source of error would inevitably increase the other, hence it becomes particularly hard to produce a model that both captures all important regularities and also generalize well. Bias refers to error from ill assumptions in the algorithms. High bias could cause algorithm fail to capture important features in training set and results in underfitting. Whereas variance refers to error from sensitivity to minor fluctuation in training set, causing model to learn from noise data and overfit.

Outlier detection is an effective way of preventing neural network from modelling features in noise, which leads to high variance, by introducing a certain level of bias. Over the decades, many methods have been proposed within the Bayesian framework. Level of anomalies is then measured through probability distribution [2].

Bimodal Distribution Removal (BDR) shows that along with epochs of training, frequency of prediction error against training set input falls into a bimodal distribution, with one mode representing maturely learned pattern and the other being outlier pattern [3]. The cause of a bimodal distribution could be mixture of two normal distribution, with mean of each distribution intuitively locate around one of the modes in bimodal distribution [4]. A mix of two normal distributions affirms the existence of outliers, which could belong to another population than normal input patterns. However, it is not uncommon that bimodal distribution arises from a number of other mechanisms, in which case deducing two populations from observation of bimodal distribution and removing one of the two modes could hinder neural network from learning all important features [5].

This paper will examine the effectiveness of ANN in classifying tumor malignancy given input cell patterns. Particularly, the efficacy of BDR in removing outlier patterns and GA's capability of filtering off noisy feature will be tested to determine if they contribute to ANN's performance.

## 2 Method

### 2.1 Data Set

In this experiment we will use Wisconsin Breast Cancer Diagnosis dataset for our examination. Benefits of choosing this dataset is many-fold. First of all, this is a relatively well-studied dataset, comparing our results with earlier best-in-kind predication models are hence made convenient. Secondly, the dataset contains more features than many other comparable datasets pertaining to this research topic, which gives us more space to prove and reject the efficacy of genetic algorithm in filtering off noisy features.

The dataset contains 569 records of cell patterns, in which 212 of them are associated with benign tumors and 357 are proved to be malignant. 30 features of each pattern are recorded and are in numerical form but with slightly different magnitude. One column of categorical data represents benign or malignant. We have filtered off the first column that contains identifiers which is not relevant to our classification mission.

### 2.2 Neural Network Architecture

Our experiment would train three ANN models that all learn from training set and try to predict against test set. All networks would maintain same number of neurons and learning algorithm, with the only difference being whether BDR for outlier pattern removal and GA for feature selection have been applied.

Hence, for ease of identification, we would call the ANN without BDR or GA the control group, the model with only BDR applied the experiment group B, the model with GA for feature selection the experiment group G respectively. Prediction results from all ANNs are collected to draw a comparison to examine the performance difference brought by BDR and GA.

As shown in Fig.2-1, the design of our model could be split into three parts: 1) pre-processing of data, converting multiple magnitude numerical features into data of same interval. 2) feature selection, implemented through genetic algorithm that will output a series of masks denoting noisy features that should be excluded from classification. 3) neural network for the actual training and classifying given selected data, and iterative BDR process that periodically triggers to detect and remove outliers from input.

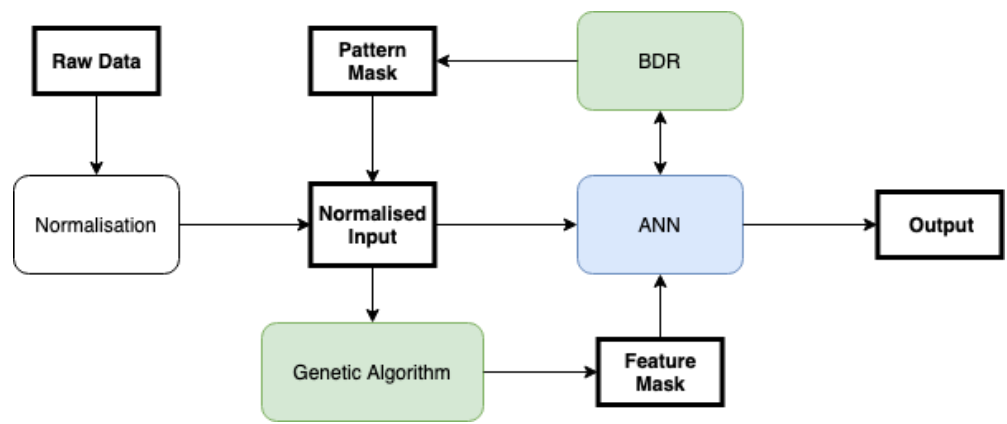

**Fig. 2-1.** High-level Architecture of Model

We have devised a 3-layered neural network consists of 30 inputs and 2 outputs. Number of hidden units is determined based on rule of thumb through a round of testing. Predicted output is the class of the output unit with highest value from Sigmoid function, normalized to binary value to remain comparable with target. Training in all experiment groups and control group ANNs is conducted for 500 epochs.

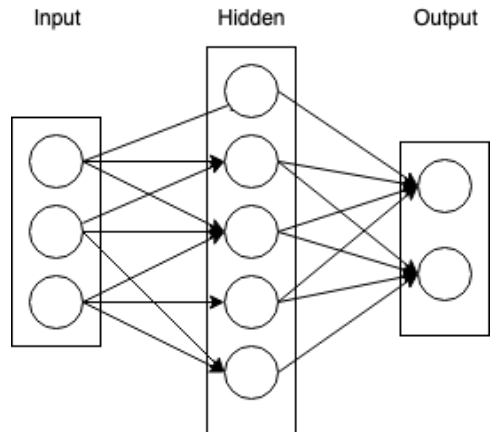

**Fig**. **2-2.** 3-Layed Neural Network

**2.3 Data Pre-Processing**

The original dataset contains 32 columns with one column for identifiers and one for result. We excluded identifier information as they often do not supply any meaningful information to the classification problem and may prolong training process. Result column is encoded in categorical form of M and B denoting malignancy and benign. We converted M for 1 and B for 0 to reserve the 2-class output.

Remaining 30 feature columns are numerical but vary in magnitude. Sola and Sevilla points out that proper normalisation of numerical data to reduce their interval difference could help reduce training time [17]. In our experiments, we use PyTorch's built in normalize function to force numerical input value into a range between 0 to 1 based on normal distribution.

**2.4 Genetic Algorithm for Feature Selection**

Genetic algorithm developed by John Holland borrows the notion from Darwinian theories of species evolution. It simulates the main steps in species evolution such as reproduction, mutation and crossover [8]. Genetic algorithm contains five basic steps: 1) coding of variable, 2) initialisation of population, 3) fitness test, 4) reproduction and 5) mutation [9].

In genetic algorithm, one use gene to denote a unique variable and chromosome to denote a possible assignment of all variables [9]. The process of the algorithm is basically search through all possible combination of gene and determine the best possible solution through reproduction and mutation. Particularly, the fitness test function determines how suitable a chromosome is in the environment and ultimately determines which solution should survive [10].

Particularly, for feature selection each gene then uniquely represents whether a feature should be selected or note. Hence, chromosome is then a binary string with length equals to input size, denoting a combination of all feature that may be selected or masked. Fitness test function evaluates how strong a set of features contribute to the target. Naturally, the fitness test function is thus a classifier that approximates the real function maps features to target. In our experiment, we will use a supporting vector machine algorithm as fitness test function, majorly due to it speed and low operational cost.

**2.5 Bimodal Distribution Removal Process**

BDR proposed by Slade and Gedeon exploits neural network's capability of distinguishing potential noise mentioned above. Two thresholds are chosen for selecting patterns to be removed. It is obvious that error frequency bimodal distribution generated by a trained neural network will mean skewed towards the mode with lower value, as an effectively trained ANN should to approximate target for most of input patterns. Hence, BDR use the mean of error distribution as first threshold and would sweep all input patterns whose predict error $e_i$ higher than $mean(E)$ into a candidate collection.

The candidates would have error set $E_c$ from which we would select the second threshold. Mean and standard deviation are calculated from the set. Again, one could infer that due to the existence of peak, mean of the candidate set will be skewed towards the right. Hence, error higher this mean could be a good flag for outlier. In practice, BDR use mean plus standard deviation as the second threshold. Input patterns whose error higher than the second threshold would be removed [3].

$$e_i - mean(E_c) > std(E_c) \qquad (2)$$

In normal distribution, data points that one standard away from sample mean are normally considered as noise. Hence, BDR process's choice of one sigma as removal space appear to be reasonable.

## 2.6 Experiment Setup

Training of neural networks are conducted with PyTorch. We split the original 569 rows of Wisconsin breast cancer diagnosis data into training set and testing set. Training set consists of 20% of the original data set, whereas testing set takes up the rest 80%.

We first train the control group model to obtain a 90% accuracy rate as the baseline of acceptable level performance estimator for GA and BDR. The same parameters (hidden units, learning algorithm, learning rate) would be applied to the experiment models. That shall enable us to isolate performance differences attribute to the usage of GA and BDR process.

### 2.6.1 Bimodal Distribution

Naturally, BDR process would turn out to be beneficial under the premise that error patterns between prediction and target conforms to a bimodal distribution after enough epochs of training. Error frequency distribution is collected for Epoch 1 and Epoch 500 in control group.

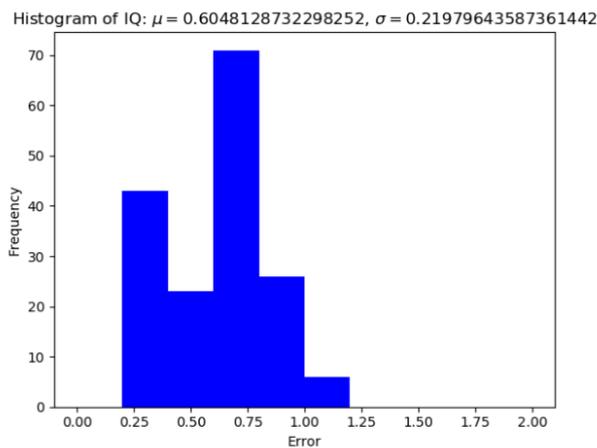

**Fig. 2-3.** At epoch 1, training set prediction error forms an approximately unimodal distribution. This is intuitively due to randomly assigned weights and our model has learned nothing from input features, predictions are largely random.

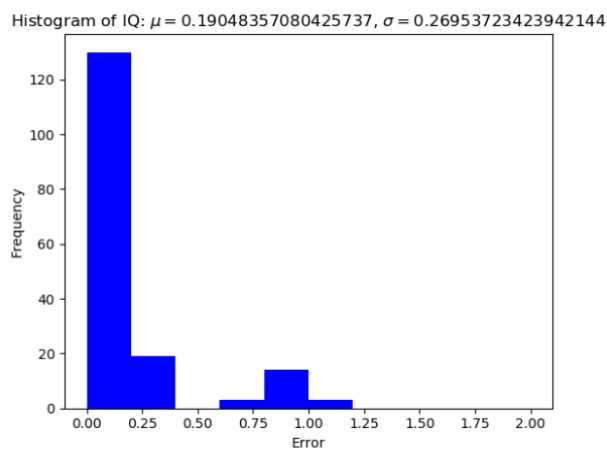

**Fig. 2-4.** At epoch 500, a bimodal distribution of frequency error could be observed. Peak at the left represents patterns that has been learned by the model, featuring high prediction match, whereas peak on the right consists of patterns that are learned even after 500 rounds of training, suggesting high possibility of being outliers.

It could be observed from Fig 2-3 and Fig 2-4 that after 500 epochs of training, the error distribution shifted from an approximately single mode distribution to a bimodal distribution. It is evident that our control group neural network is learning the features, hence in Epoch 500 most of error patterns fall within the range of 0 to 0.1. Furthermore, the mode centered around 0.75 to 1.00 in Figure 2-4 represents training set input that are hardly learned by our model, and hence should be reasonably treated as candidates for BDR. Our observation in control group conforms to what presented by Slade and Gedeon in their original paper of BDR [3].

It should be worth noting that although as mentioned before, predict output represents the class of output neuron, error data in plotting error frequency distribution are calculated based are the unnormalized highest activation output. Particularly, error is calculated as the absolute value of target subtracting highest activation output.

$$e_i = t_i - Sigmoid(z_i) \tag{1}$$

**2.6.2 Loss Function and Evaluation Function**

Loss function in feed-forward neural network calculates the difference between target and prediction, based on which optimization could be done to adjust connection weights for better approximation of the true function from feature to target value. In our classification experiments we use cross-entropy as loss function majorly for its performance.

Evaluation is based on percentage of accuracy comparing predictions generated by our ANNs and the actual target results.

**2.6.3 Hyper Parameters and Optimizer**

The control group neural network is built with 30 input neurons that each correspond to one feature and 2 output neurons that map to 2 classes of classification output. We start out testing with 40 hidden units and Stochastic Gradient Descent (SGD) and quickly discovered it is hard to achieve testing set accuracy higher than 70% with any configuration of learning rate. In fact, with SGD plus momentum most of time we end up with producing around 30% accuracy, suggesting significant difficulty with SGD to escape from local optimal. Adam on the other hand was able to produce stable testing set prediction accuracy of higher than 90%. With our dataset, 0.01 learning for Adam seems to work the best, as higher learning rate incurs much more significant oscillation in loss curve, which results in less ideal predication result.

In addition, with the chosen optimizer we noticed that loss curve over training epochs often reaches its bottom at around 400 epochs, further training in fact incurs loss increasing as shown in Fig 2-5. We repeated test with 400 and 500 epochs rounds of training for 30 times respectively and average predication accuracy yielded accuracy rate of 91.6% and 90.5% respectively. It indeed shows 400 epochs generates more optimal training results.

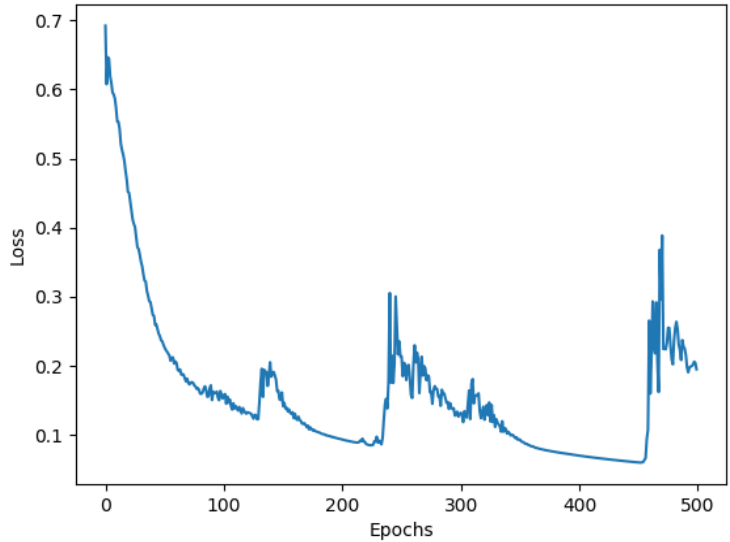

**Fig. 2-5**. Loss over epochs under Adam, learning rate = 0.01

We then set off to choose the number of hidden units that strikes the balance between adequate accuracy and avoiding over complex network structure. Tests were run against various configuration of hidden unit numbers. Following the rule of thumb, we start with 24 hidden units and increase 2 at each time. The aggregated results are presented in Table 2-1. Although results yields are in very close range, 40 hidden units appear to be the best configuration with our dataset.

| Hidden Units | Accuracy |
| --- | --- |
| 24 | 0.91578947 |
| 26 | 0.91085526 |
| 28 | 0.9122076 |
| 30 | 0.91480263 |
| 32 | 0.9145614 |
| 36 | 0.91582602 |
| 38 | 0.91654135 |
| **40** | **0.91672149** |
| 44 | 0.91306043 |
| 48 | 0.91355263 |
| 52 | 0.91212121 |
| 56 | 0.9120614 |

**Table. 2-1.** Testing set accuracy obtained by various configuration of hidden units

**2.6.4 Experiment Groups Configuration**

Separate exercises are conducted to train the experiment groups. A same 3-layered neural network with 40 hidden units is trained using Adam with 0.01 learning rate for 400 epochs. For experiment group G, genetic algorithm will run before the training of neural network as feature masks need to be produced and applied before information is fed into ANN. We configure the population size to be 20, crossover rate 0.2 and allow 2 genes to mutate in each round to allow sufficient mutation from happening. Genetic algorithm will run for 10 generations for each experiment, the produced chromosome with best fitness value after all generations are then decoded to mask features that are selected. ANN is then trained the same as in control group, but only selected features will be taken into account.

For experiment group B, outlier detection process will run after every 50 epochs and remove identified outliers from input patterns. Training will then continue with updated input data set. BDR process would normally require a termination point, post which removal will no longer be in action. If removal process is not stopped it would eventually remove inputs to an excessive extend, resulting in less than sufficient information being feed to the model. Hence, one should either stop BDR process once aggregated loss fall under a threshold or actively determine if a bimodal distribution of prediction error is no longer present.

Prediction result of both experiment group models are then compared with results obtained by control group model. For experiment group G, we expect to observe higher accuracy rate as it should prune features that are noisy or irrelevant. We also expect to observe smaller neural network size and faster training time as a result of less input. Should we fail to observe these performance improvements, we would need question our implementation of genetic algorithm, that may be supporting vector machine is not an ideal choice as fitness test function.

If BDR is as effective as it claims, we should expect higher accuracy against testing data from experiment group B, as removal of outliers avoids model to learn from noise and hence should exhibit better capability of generalization. Another observation we expect, although trivial, is the reduction of training epochs and higher accuracy against training set, as now there are less information, which are all supposedly quick to model as they are noise-free, supplied to the model. However, should these expected observations not emerge e.g. experiment group B shows lower accuracy rate, it may be indicated that patterns removed by BDR process are not truly noise data but meaningful patterns. Removing non-noise input will naturally lead degraded performance in generalizing. Furthermore, it will reject the hypothesis that bimodal distribution is formed due to input data are sampled from two populations, instead some other mechanisms lead to the formation of two modes.

# 3  Result and Discussion

Evaluation results are gathered for the control group model and all three experiment group models. We will compare the results obtained pair-wise to determine: 1) What is the impact of GA-based feature selection and how does it impact neural network's classification performance? 2) Whether BDR effectively removes outlier pattern or not, do these removals improves predictions accuracy or reduces training time? And more fundamentally, does observation of bimodal distribution necessarily supports the existence of outliers?

As mentioned in previous section, we are able to obtain prediction accuracy in control group with 40 hidden units, Adam with 0.01 learning rate over 400 epochs of training

**3.1 Effectiveness of GA-based Feature Selection**

To assert the impact of feature selection on classification results, we compare results obtained from experiment group G to the control group. We set cross over rate to 0.2 and number of mutations to 2 genes per round to allow sufficient mutations happens, though these configurations ultimately only indirectly contributes to classification accuracy through feature mask.

With all other hyper parameters held the same, we train and test experiment group G for 10 times and yield average accuracy rate of 92.3%, a non-trial improvement comparing to control group's 91.6% average result. However, despite the better average accuracy, we do notice certain test rounds produce much lower accuracy rate of about 89%. Other than random initial weights, there may be some other factors contributing such significant variance. Hence, we repeat training and evaluation process for experiment group G for 400 times, recording number of features remaining after GA-based selection and accuracy to explore if any correlation exists.

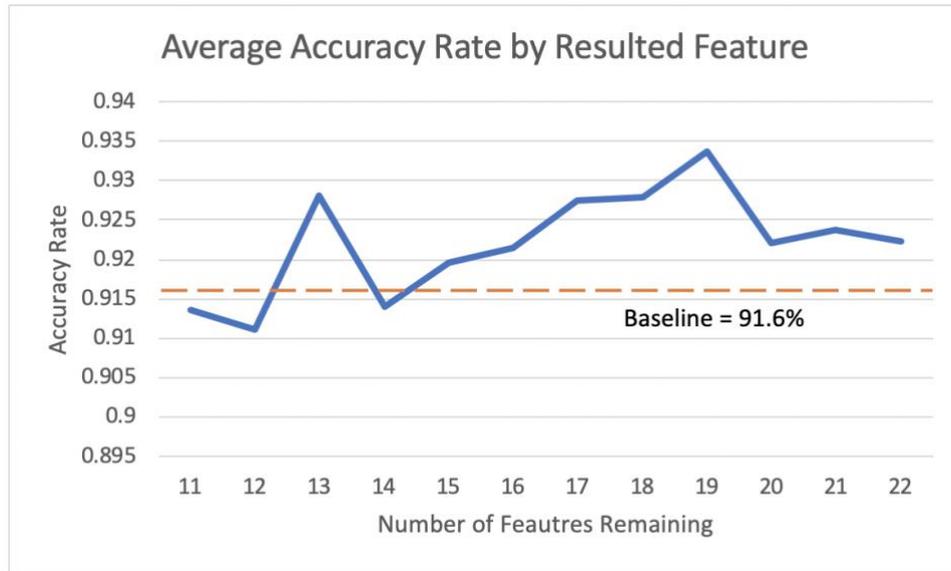

**Fig. 3-1**. Testing set accuracy from 400 rounds of experiment are collected and grouped by number of remaining features after GA selection mask, average accuracy rate of each group is plotted. Group of data points less than 10 are excluded from the plot

As shown in Fig 3-1, best accuracy rate is obtained when number of remaining features is between 16 to 20, with an exception of 13. Lower accuracy in experiment groups of less feature may be an indicator of excessive pruning, resulting in insufficient training of neural network, whereas inferior performance in groups of more than 20 features could be factored to inadequate pruning, hence noised features are accounted for during training.

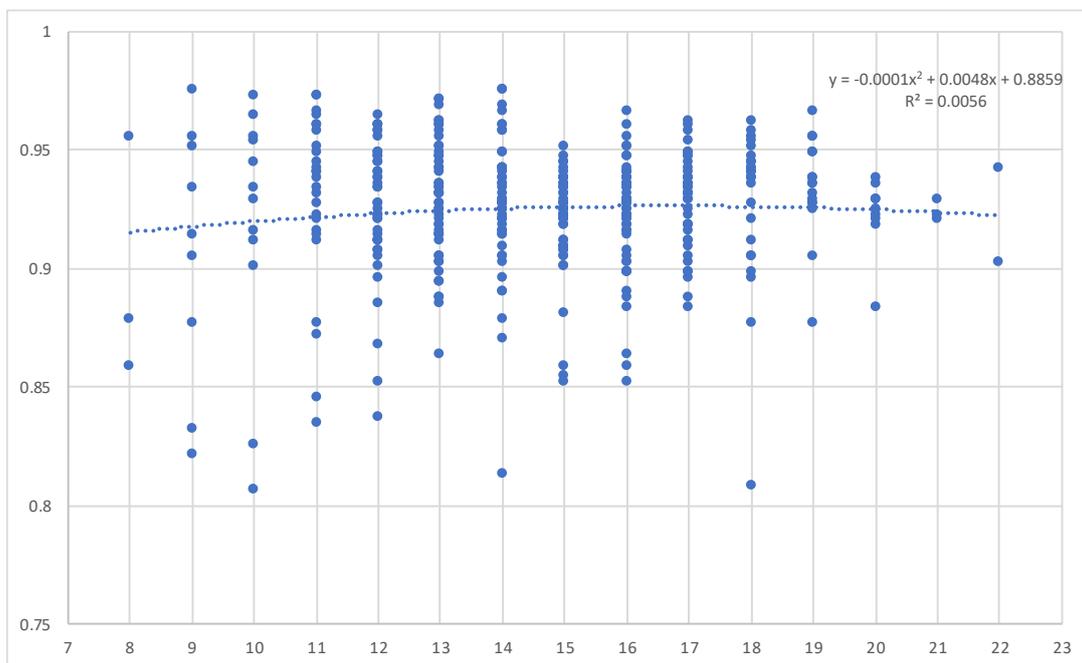

**Fig. 3-2**. 400 experiment results are plotted, with a non-linear regression equation inferred with coefficient of determination presented

However, as presented in Fig 3-2, there is no clear correlation between number of features remaining after GA-based selection and test accuracy. Coefficient of determination measures how much percent of change in one sample could be explained by change in another sample. And in our case, only 0.56% of change in accuracy could be explained by number of features, which is far less than significant. Thus, relationship between remaining features and test accuracy appear to have more murkiness. Though experience gained from previous experiments points out to banding resulted features in between 16 to 20, which produces best results, such effective banding is anticipated to change when a different dataset or different architecture of neural network model is applied. Hence, how to ensure GA-based selection produce best improvements to neural network and being able to generalise such optimization remains a challenge.

Another angle to measure whether GA-bases selection has improved neural network performance is by looking at if the number of hidden units required to produce baseline prediction accuracy could be reduced.

To examine if less hidden units are needed to produce adequate classification accuracy, instead of fixing number of hidden units at 40, we will dynamically calculate it based on number of features remaining after applying feature mask from GA selection. We create a series of increment bands from 0 to 24, interval by 2. The actual number of hidden units will be calculated by number of remaining features plus the assigned increment band.

Hence, given any number of remaining features we produce 12 sets of hidden units that could be used for experiment. Effectively, we are trying to examine what are the increment bands based on remaining features could produce accuracy at least same as base line.

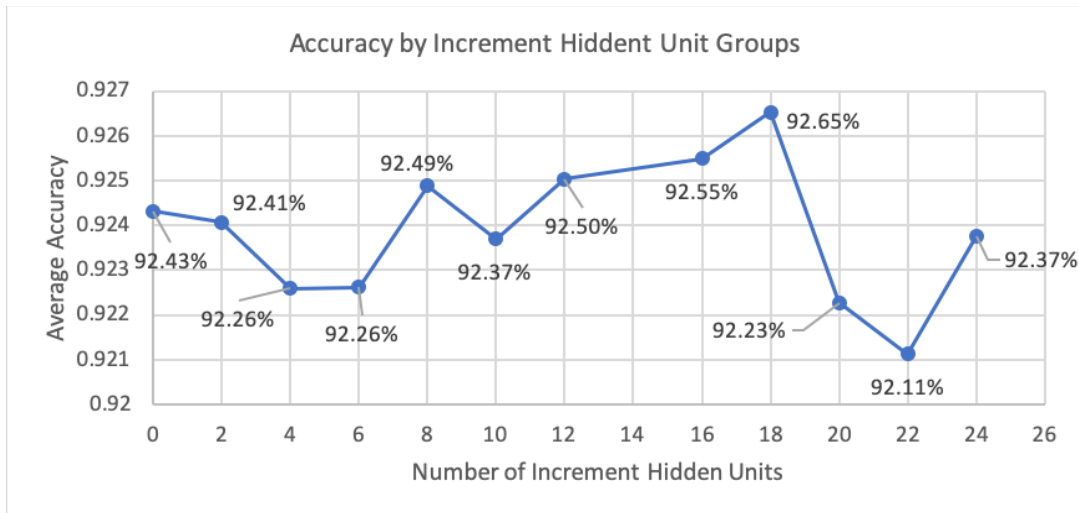

**Fig. 3-3.** Results from total 1200 rounds of training are plotted, grouped by their increment band. Actual number of hidden units = size of remaining features + increment band

We conducted 100 rounds of GA-selection, each resulting set of pruned features are fed into the series of increment bands to determine the actual number of hidden units. Hence, a total number of 1200 of neural networks are trained with various feature vectors and different number of hidden units. Results are presented in Fig 3-3.

The result is surprisingly good, with nearly all the increment bands producing accuracies higher than 91.6% baseline. Particularly, increment of 12 to 18 neurons based on input size appear to yield best average results. Consider the most effective features groups are between 16 – 20, at minimum we only need 28 (16 input size plus 12 increment) number of hidden units to produce good enough prediction accuracy. This is a significant reduction on size of hidden layer comparing to 40 hidden units required to produce baseline result in the control group.

**3.2 Effectiveness of Bimodal Distribution Removal**

Training on experiment group B was conducted with same hyper-parameters as the control group. As variance threshold as the termination point and sigma distance used to decide outlier candidates both affects the pruning process of input patterns. We control these two variables and construct experiment cases for each combination of them. Aggregated results based on 30 neural networks separately trained and tested for every experiment case are presented in Table 3-1.

| Variance Threshold | Sigma Distance | Epochs | Remaining Input Size | Accuracy |
|---|---|---|---|---|
| **0.01** | 1 | 400 | 94.50 | 87.1711% |
| | 1.1 | 400 | 98.85 | 86.4035% |
| | 1.2 | 400 | 98.37 | 87.1564% |
| | 1.3 | 400 | 99.23 | 87.7248% |
| | 1.4 | 400 | 100.70 | 88.2018% |
| | 1.5 | 400 | 101.88 | 88.3553% |
| | 1.6 | 400 | 102.63 | 87.9449% |
| | 1.7 | 400 | 103.24 | 87.6563% |
| | 1.8 | 400 | 104.00 | 87.1686% |
| | 1.9 | 400 | 104.50 | 87.3991% |
|      | 2   | 400 | 105.01 | 87.8130% |
|      | 1   | 400 | 90.30  | 86.4254% |
|      | 1.1 | 400 | 95.20  | 87.3575% |
|      | 1.2 | 400 | 98.20  | 88.1067% |
|      | 1.3 | 400 | 98.93  | 88.3004% |
|      | 1.4 | 400 | 100.82 | 88.7851% |
| **0.05** | 1.5 | 400 | 101.83 | 88.8012% |
|      | 1.6 | 400 | 102.50 | 88.9442% |
|      | 1.7 | 400 | 103.08 | 88.9419% |
|      | 1.8 | 400 | 103.80 | 89.2373% |
|      | 1.9 | 400 | 104.34 | 89.1974% |
|      | 2   | 400 | 106.80 | 89.3760% |

**Table. 3-1.** Experiment Group Result

Variance threshold in the first column serves as a termination point, beyond which we will stop training epochs to avoid excessive deletion of input patterns. Sigma distance in second column represents how many standard deviations away from $mean(E_c)$ should a pattern's error become that we would categorize an input pattern as outlier. It is worth pointing out that variance here refers to the stand variance of the classification error distribution. Epochs and remaining input size are respectively the last epoch and number of patterns retained in training set when loss reduced to target variance threshold.

Results from experiment group B seems to reject our hypothesis that BDR process shall effectively removes outliers in training set and thus present better prediction result against test set. From Table 3-1 however, none of the configuration outperforms the control group model (which exhibits stable >91.6% accuracy against test set). The closest configuration of experiment group model compared to the baseline is the one with 2 sigma distance and variance threshold 0.05, obtained average 89.37% accuracy. However, the retained 106 patterns (originally 113) suggests that the input patterns are barely modified and still, it still produces results that worse than the baseline.

### 3.2.1 Overfitting

To understand if BDR process caused our model be overfitting, hence perform less than ideal in testing sets, we produce the loss curve from both training data and testing data during each epoch of training. If test set loss curve begins to increase after certain point along with training, it may be an indication of overfitting.

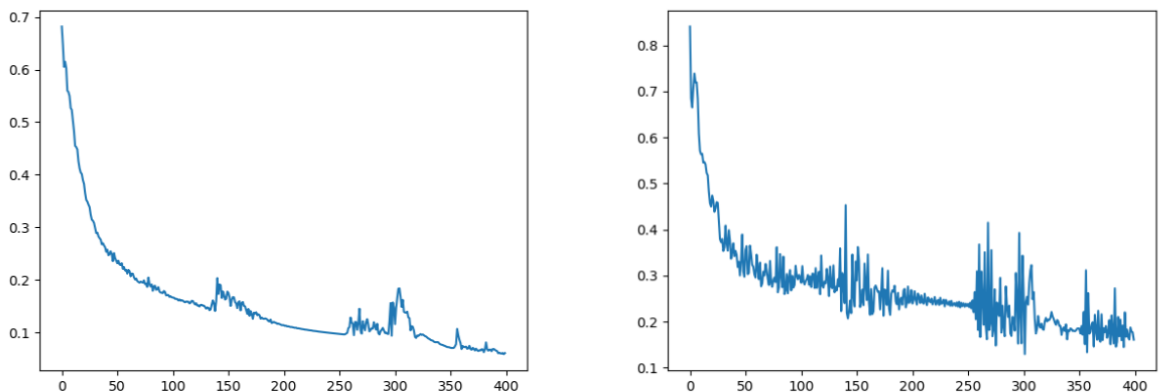

**Fig**. **3-4.** Training set loss curve by epoch on the left, test set loss curve on the right.

However, from data presented in Fig 3-4, loss curve on test set does have bigger oscillation, but still in general decreases along the time. It is insufficient to state that less ideal performance in experiment group B is due to overfitting.

### 3.2.2 Ineffective Termination Point and Non-existence of Premise

It could be further observed that test accuracy is better when remaining input size is larger, which means those experiments with less patterns removed by BDR produce better performance. To formalize this observation, we run a linear regression on data presented in Table 3-1, aggregating those produced under variance threshold 0.05 and 0.01, and we yield positive correlation between size of remaining pattern and test accuracy as shown in Figure 3-5.

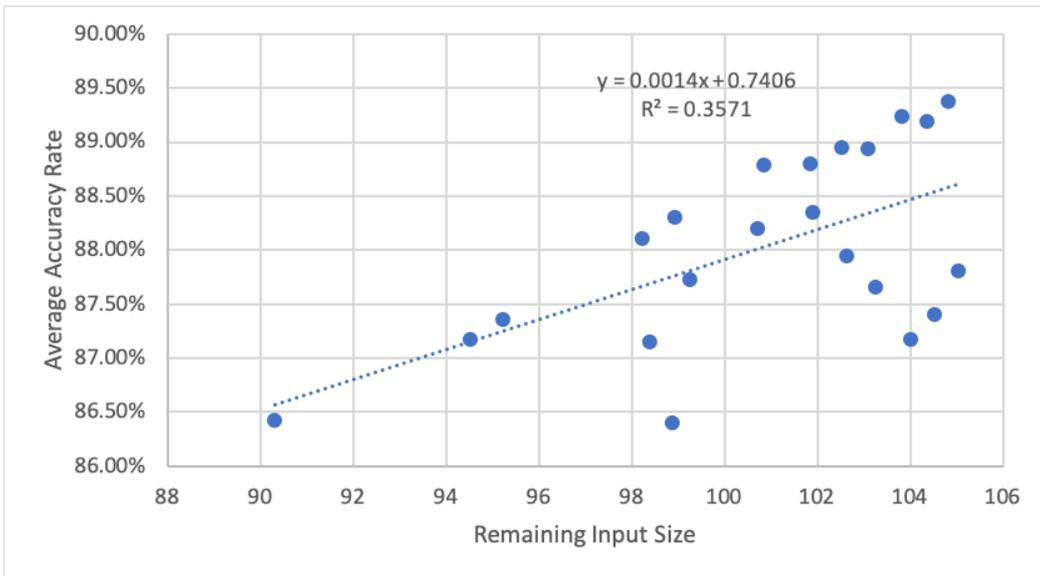

**Fig. 3-5.** Correlation between Remaining Input Size and Test Accuracy

We obtained a significantly strong positive correlation coefficient of 0.0014, suggesting increasement of 1 in input size would add up 0.14% accuracy rate. We also have coefficient of determination $r^2$ of 0.357, suggesting 35.7% of changes in the sample of accuracy rate could be explained by change in input size.

At this point, we've rejected the hypothesis that in this classification problem, BDR process could improve generalization and thus yield better test result. We further identified that the less patterns removed by BDR process, the better prediction results approximate the target. It appears that, BDR implemented in our model does not effectively identify outlier. Patterns removed by BDR process in our experiment group model are then largely valid input features, removal of which naturally results in downgrading of prediction. Although our experiment is not enough to invalidate BDR's effectiveness in general, it is certainly valuable to explore the possible causes that BDR do not yield performance improvement as it had demonstrated in other research works [3].

One explanation questions the effectiveness of prediction error variance as the termination point of whether all (or enough) outliers have been removed from input patterns. It should be the case, an arbitrarily chosen variance threshold may allow adequate outliers to be removed but could also encourage BDR process to excessively delete patterns from input set. Instead, a more reasonable termination point for BDR should be when bimodal distribution is no longer present. Dip test of unimodality proposed by Hartigan and Hartigan could applied before each round of BDR process to determine if outliers have been effective removed. Dip test examines whether a distribution is unimodal or multimodal [7]. When the frequency error distribution is unimodal, it could be inferred that most of patterns are effectively learned whereas remaining "slow coaches" could be learned but more training time should be allowed [3].

Another possible explanation rather challenges the premise of BDR exists in our experiment. It could be the case that the mixture of two unimodal distributions (like normal distribution) which hypothetically caused the formation of bimodal distribution does not really exist. It is rather some other mechanisms that attribute to this bimodal distribution. Hence, bimodal distribution itself is no longer a valid indicator of outliers in this case.

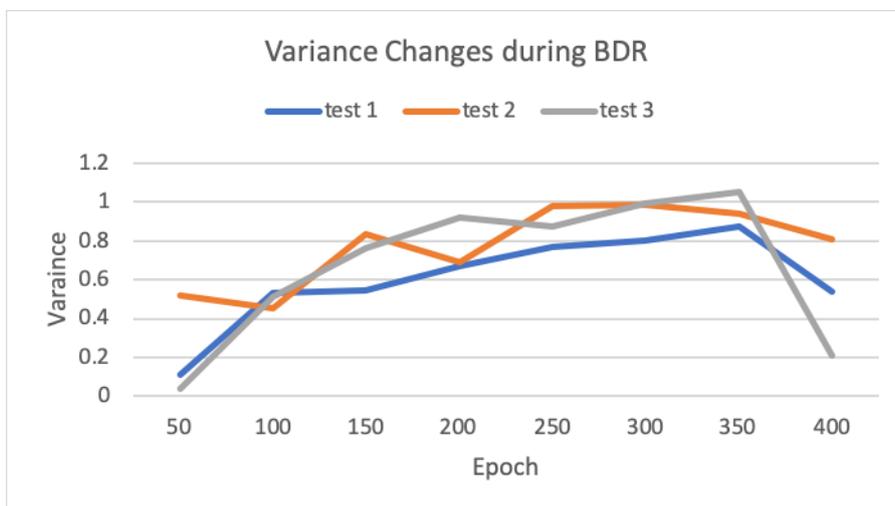

**Fig. 3-6.** Change of variance long with patterns removed by BDR throughout training epochs.

Although it is hard to examine the existence of two overlapping distributions, assuming its existence and prove by contradiction is not improbable. If the premise of BDR's efficacy holds, patterns removed by BDR process should be or having good possibility to be outliers, then variance in predication error distribution should decrease along with epochs increase. Fig 3-6 shows results gathered from our experiment with three different configuration of variance threshold and sigma distance used by BDR, variance from prediction error distribution is plotted against epochs in which BDR is triggered. However, among all three groups, none of them shows a constant decreasing trend of variance along with training. Instead, variance increases during the training process, producing a contradiction to the assumption that two overlapping unimodal distribution exists in the prediction error distribution.

## 4   Conclusion and Future Work

This paper explores the effectiveness of a GA-based feature selection approach and Bimodal Distribution Removal method in improving generalization performance of neural network in classification problem. Particularly, we explored whether GA has effectively improved classification accuracy and reduced network size. We also focus on BDR's theoretical premise that bimodal distribution of training error could attribute to the mixture of two unimodal distribution. Also, we explore the choice of termination point in BDR process that generates the best performance.

From experiment group G we discovered that: 1) GA-based feature selection does improve classification accuracy and could effectively reduce result network size. 2) It is still a challenge on how to bound selection process to produce best feature mask results. Future works could explore if a fitness test function other supporting vector machine could yield better results, or if there are mechanisms to reduce variance in performance of selected features.

From experiment group B we have observed that: 1) BDR does not yield better performance in prediction against test set comparing to baseline and this could be due to the premise of BDR, that the prediction error distribution is a mixture of two unimodal distribution, does not hold. 2) Predication loss may not be a good termination point for BDR as it could result in excessive deletion.

As discussed in Section 3, we hypotheses that using the dip test to determine if bimodal distribution still exists as termination mechanism instead of applying prediction loss as termination threshold could potentially improve prediction performance. Though we have not incorporated the test in this paper's experiment, future researcher could certainly explore if our hypothesis holds. Furthermore, although we have presented an inference in the section by which we state there is a contradiction to the existence of BDR's premise in our experiment setup, it is still not clearly known that how to effectively determine whether a bimodal distribution is due to a mixture of two unimodal distributions. Future researchers could certainly look into this area.

The advantage of Bimodal Distribution Removal and other earlier outlier detection methods like Least Trimmed Squares centralize at their simplicity in implementation and comparably great performance improvement. Outlier detection introduce bias into the model in hoping that variances are fairly reduced. However, fine-tuning these methods to avoid a too strong bias being introduced remains challenging to certain problems. In data sets where detection of outliers is difficult or less beneficial than expected, researchers could consider alternative school of methods like pruning and cascade network to make neural network robust to variance.